\documentclass[letterpaper, 10 pt, conference]{ieeeconf}  

\IEEEoverridecommandlockouts                              
\usepackage[utf8]{inputenc}
\usepackage[T1]{fontenc}
\usepackage{amsmath}
\usepackage{graphicx}
\usepackage{cite}
\usepackage{lmodern}        
\usepackage[T1]{fontenc}    
\usepackage{textcomp}       
\usepackage{mathptmx}  
\usepackage{newtxtext,newtxmath}

\pdfmapfile{=pdftex.map}

\title{\LARGE \bf
Physical Reservoir Computing in Hook-Shaped Rover Wheel Spokes for Real-Time Terrain Identification}

\author{{Xiao Jin$^{1}$, Zihan Wang$^{1}$, Zhenhua Yu$^{2}$, Changrak Choi$^{3}$, Kalind Carpenter$^{3}$, Thrishantha Nanayakkara$^{1}$
\thanks{*This research is partially supported by the European Union’s Horizon 2020 Research and Innovation Programme under Grant Agreement No. 101016970 (Natural Intelligence).}
\thanks{$^{1}$ Xiao Jin, Zihan Wang, and Thrishantha Nanayakkara are with Dyson School of Design Engineering, Imperial College London, SW7 2DB, London, UK.
        {\tt\small e-mail: xj521@ia.ac.uk}}%
\thanks{$^{2}$ Zhenhua Yu is with the Department of Computer Science, University of Aberdeen, AB24 3UE, Aberdeen, UK.
        {\tt\small e-mail: zhenhua.yu@abdn.ac.uk}}%
\thanks{$^{3}$ Kalind Carpenter and Changrak Choi are with the NASA Jet Propulsion Laboratory, 4800 Oak Grove Drive, La Cañada Flintridge, CA 91011, USA.
        {\tt\small kalind.carpenter@jpl.nasa.gov}, 
        {\tt\small changrak.choi@jpl.nasa.gov}}%
}
}

\begin{document}
\maketitle

\thispagestyle{empty}
\pagestyle{empty}

\begin{abstract}

Effective terrain detection in unknown environments is crucial for safe and efficient robotic navigation. Traditional methods often rely on computationally intensive data processing, requiring extensive onboard computational capacity and limiting real-time performance for rovers.  This study presents a novel approach that combines physical reservoir computing with piezoelectric sensors embedded in rover wheel spokes for real-time terrain identification.  By leveraging wheel dynamics, terrain-induced vibrations are transformed into high-dimensional features for machine learning-based classification.  Experimental results show that strategically placing three sensors on the wheel spokes achieves 90$\%$ classification accuracy, which demonstrates the accuracy and feasibility of the proposed method.  The experiment results also showed that the system can effectively distinguish known terrains and identify unknown terrains by analyzing their similarity to learned categories.  This method provides a robust, low-power framework for real-time terrain classification and roughness estimation in unstructured environments, enhancing rover autonomy and adaptability.

\textit{Index Terms}—Exclusive Senser, reservoir computing, Real-Time terrain identification.
\end{abstract}

\section{INTRODUCTION}

As mobile robots increasingly operate in unknown and complex environments, effective terrain detection has become essential to support autonomous exploration in fields \cite{111} like NASA space missions\cite{4161571}, disaster response\cite{9220149}, and environmental monitoring\cite{6161683}. These robots must navigate a diverse array of terrains while reliably identifying and characterizing the ground they traverse to avoid hazards and ensure mission success\cite{9200702}. In planetary exploration or rescue operations, the ability to accurately sense and interpret the terrain environment is crucial for robots to operate autonomously, adjust to changing conditions, and make real-time decisions without human intervention. This capability allows mobile robots to complete high-stakes tasks, such as collecting scientific samples\cite{8396726}, mapping dangerous zones or assisting in search-and-rescue missions\cite{8598942}.

Current advancements in smart sensor technologies have introduced various terrain detection methods that utilize such devices to meet these needs. Visual sensors\cite{10258154}, such as cameras, capture high-resolution images for analyzing surface features using deep learning and machine learning \cite{vision, audio1, v1, v2, Atha2022MultimissionTC, Sheppard2023AutomaticDP}, while lidar sensors use laser pulses to create 3D maps of the surroundings. Kurobe et al.\cite{audio1} and Zürn et al.\cite{audio2} presented the audio sensors to detect sounds associated with surface impacts or changes in combination with the camera to improve the accuracy, and  Knuthet et al. \cite{IMU1} proposed the IMUs based terrain identification method that gathers data on acceleration and orientation to help the robot understand surface texture and incline. Moreover Ahmed et al. \cite{IMU2} developed a low cost method using MEMS-IMU for localization of a land vehicle. Together, these devices enable real-time environmental analysis, providing critical data on ground conditions. When integrated with machine learning algorithms and neural networks\cite{DL}, these sensor systems can classify terrain types, predict obstacles, and improve navigation accuracy, making robots more responsive and capable in complex and dynamic environments.







\begin{figure}[ht!] 
\centering
\includegraphics[width=2.8in]{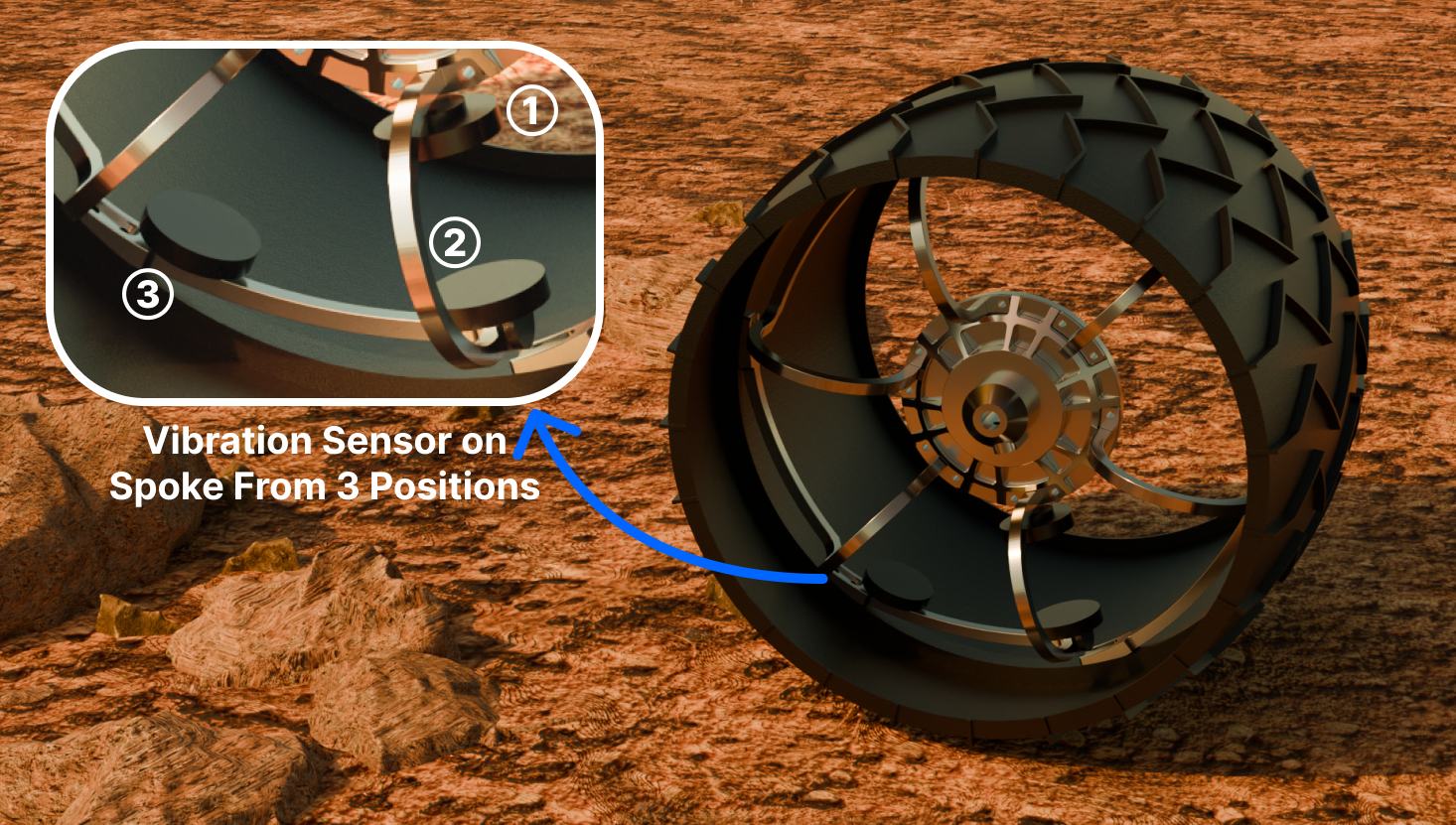}
\caption{
The hook shaped spoke in a rover wheel from NASA Jet Propulsion Laboratory. We hypothesized that this shape leads to separation of information in the stresses and their frequencies due to the geometric influence on the bending moment distribution.
}
\label{system}
\end{figure}

However, achieving high detection accuracy requires extensive real-time data processing. Each sensor generates vast amounts of data that must be continuously analyzed for rapid decision-making \cite{554205}. Visual and lidar sensors provide high-resolution terrain information \cite{lidar, muti}, but demand significant computational resources for image processing and 3D mapping, and sensitive to lightning conditions as well \cite{Cloud}. Similarly, IMUs and audio sensors contribute additional insights into surface vibrations and obstacles, but prone to noises and external disturbances that increase the complexity of terrain analysis \cite{audio2, sound}. In unstructured environments, the absence of pre-existing datasets poses further challenges for machine learning models, increasing the risk of terrain misclassification and mission failure \cite{muti}.

Currently operating NASA Mars rovers, Perseverance and Curiosity, have wheels that utilize spokes flexure with a hook-shaped profile for shock absorption (see Figures \ref{system} and \ref{harmonic}). Although terrain-specific information separation along the spoke has not been part of the original intention of this design, we hypothesized that this profile makes wheel spokes behave like a mechanical reservoir to separate vibration frequencies along the hook-shaped profile when the wheel rolls over different terrain types. 

In this paper, we test this hypothesis with a view to classify sand, rocks, or soil just using three vibration readings along the hook-shaped spoke. Physical reservoir computing utilizes natural dynamics of structures to localize unique information in a complex dynamical system. This allows us to take readouts from different locations of the reservoir to simple build regression models. 

An example is how spiders detect vibrations through leg hairs, biomimetic “spider-leg” sensors integrated into suspensions enable robots to "feel" terrain while reducing energy consumption \cite{spider, zhenhua}. While these sensors have been explored in underwater navigation \cite{underwater}, micro-robotics, and industrial inspection \cite{VibroTouch}, their application in wheel-based terrain detection remains underexplored, despite the valuable data wheel vibrations provide. Integrating this approach could improve robotic adaptability and efficiency. In this case, we can use the hook-shaped spoke as a mechanical reservoir to process terrain-induced vibrations with minimal computational cost \cite{RC}, which is of significant importance for rovers with limited computational resources. 

The rest of the paper is organized as follows. Section \ref{FEA} shows the FEA simulation results to estimate the best sensor placement locations to capture distict stress related information. Section \ref{Experimental Method} shows the basic test platform that includes six different terrains, validated the theoretical model's accuracy using the initial collected data, and established fundamental feature extraction methods based on the data. Section \ref{RESULTS} shows the Trained a model using the SVM algorithm, achieving a 90\% accuracy rate. By utilizing Euclidean distances and Mahalanobis distances, some untrained terrains were also successfully identified.





\section{FEA Simulation Based Sensor Placement Optimization}
\label{FEA}


Our approach combines Finite Element Analysis (FEA) and plug-and-play piezoelectric sensors, strategically placed on wheel spokes to capture optimal vibration signatures for terrain differentiation. Integrating piezoelectric sensors into the wheel suspension system enables precise vibration and tactile sensing for autonomous terrain recognition. 

To investigate vibrational responses and optimize sensor placement, a harmonic response analysis was conducted in ANSYS 2024 R1 on a 1:4 scaled Mars rover wheel spoke. The model was treated as a 3D solid structure, with plastic epoxy resin as the material. This analysis aimed to identify resonance frequencies that could amplify terrain-induced vibrations and guide sensor placement.

The boundary conditions were set by fully constraining the fixed end of the spoke to simulate its attachment to the wheel hub. A sinusoidal force of 3 N was applied at the free end of the spoke, representing external excitations from terrain interactions. The simulation covered a frequency range of 0-500 Hz with 50 solution intervals, ensuring a detailed frequency response. The model was meshed with 73,356 nodes and 43,335 elements, providing high-resolution stress and deformation data. The solver was configured with zero damping, isolating the intrinsic vibrational response without external energy dissipation.

\begin{figure}[ht!] 
\centering
\includegraphics[width=2.8in]{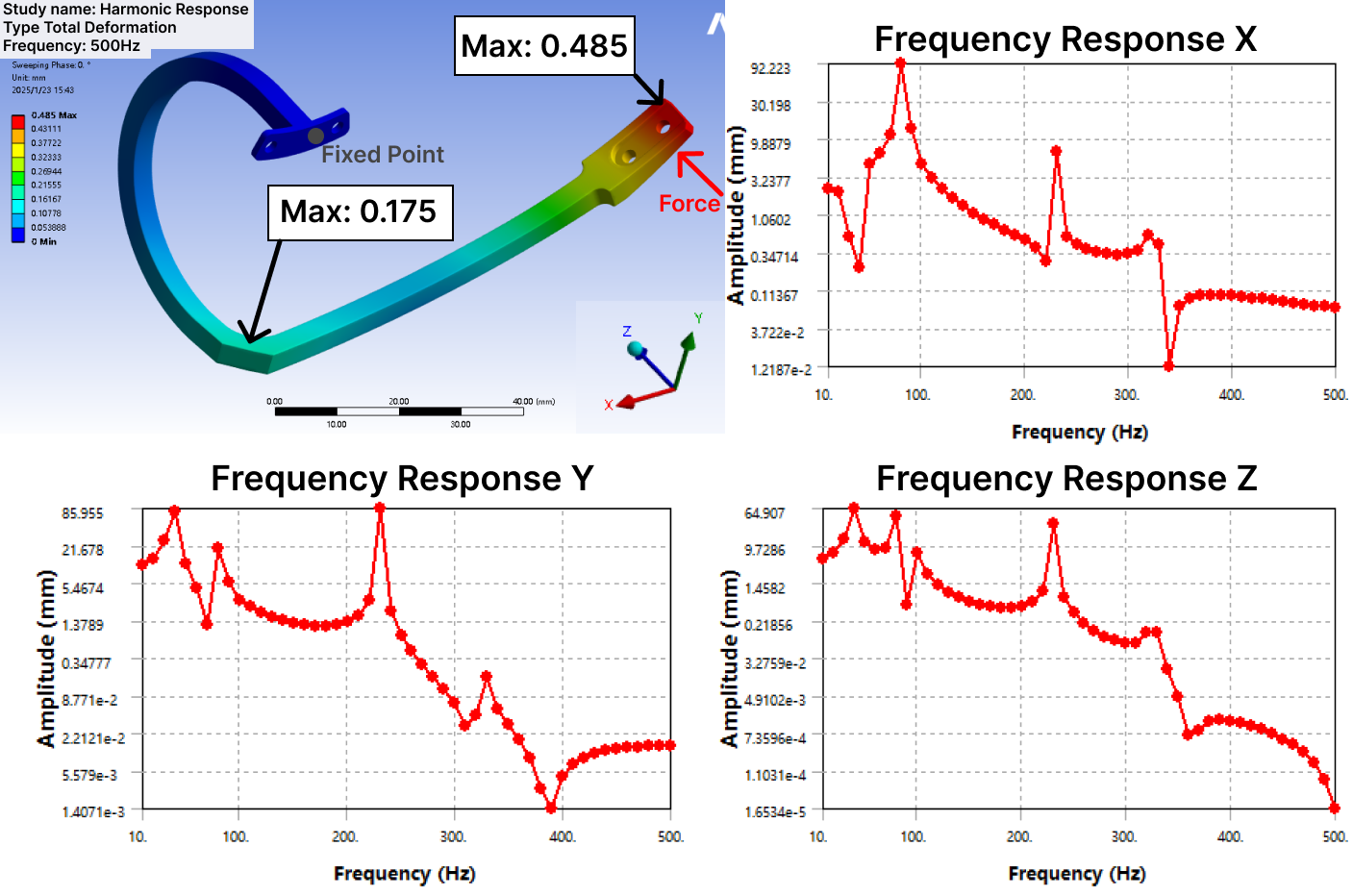}
\caption{
Harmonic response of the entire wheel spoke with 3N act on the end tip of the spoke within the 0-500 Hz range. The vibrational behavior is non-uniform, exhibiting resonance peaks at 90 Hz and 200-300 Hz, which correspond to different deformation modes across the structure.
}
\label{harmonic}
\end{figure}

The results, shown in Figure \ref{harmonic}, reveal distinct resonance peaks at 90 Hz and 200-300 Hz, corresponding to different deformation modes. Low-frequency modes (<100 Hz) result in global deformations affecting the entire spoke structure, whereas mid-to-high-frequency modes (200-300 Hz) cause localized resonance effects, leading to amplified terrain-induced vibrations.

\begin{figure*}[ht!] 
\centering
\includegraphics[width=\textwidth]{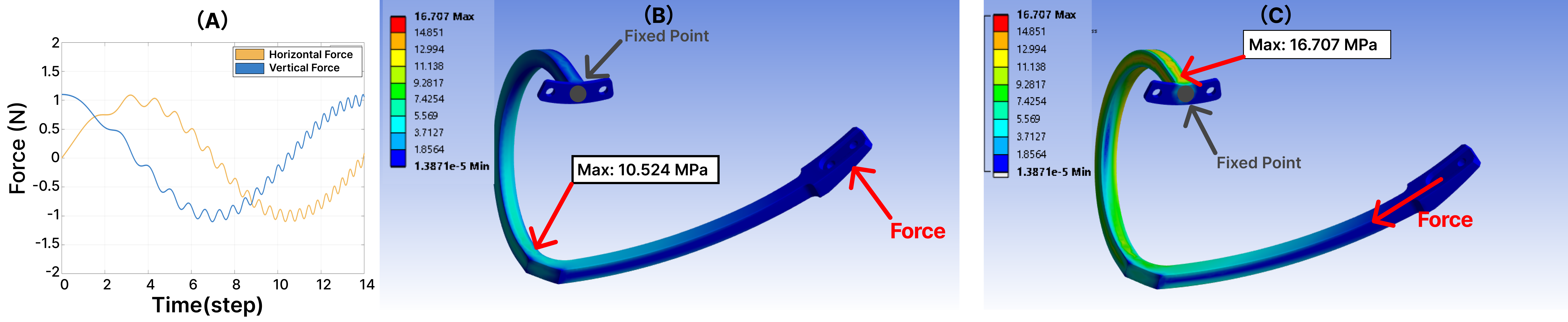}
\caption{
(A) Time series plot illustrating the variations in horizontal and vertical force components acting on the spoke over discrete time steps.  The larger oscillations reflect wheel rotation forces, while the smaller vibrations indicate terrain fluctuations; (B) Finite element analysis (FEA) results showing the equivalent stress distribution on the spoke under a primarily vertical force. The mounting holes serve as fixed constraints, with peak stress reaching 10.524 MPa at the lower curved region; (C) FEA results showing the equivalent stress on the spoke under a horizontal force. The peak stress shifts to the upper fixed point, reaching 16.707 MPa, indicating significant stress concentration.}
\label{Entropy}
\end{figure*}

High-frequency vibrations (200-300 Hz) enhance terrain texture signals, revealing that certain spoke regions naturally amplify terrain interactions. However, while these signals provide valuable terrain features, optimizing sensor placement remains a challenge. Without strategic positioning, critical terrain-specific frequencies may be overlooked, reducing classification accuracy.

To address this, we analyzed stress concentration variations alongside vibrational responses. The results show that different terrains induce distinct stress distributions across the wheel structure. This finding suggests that stress concentration patterns can act as terrain-sensitive amplification mechanisms, highlighting dynamic features that are often obscured in vibration analysis alone. Unlike pure vibration signals, stress concentration is influenced by both internal material responses and external excitations, effectively functioning as a mechanical filter that selectively amplifies terrain-specific responses.

However, a key challenge remains: while stress-based amplification improves terrain classification, its effectiveness depends on a precise understanding of how structural forces interact with vibrational modes. To investigate this, we conducted finite element analysis (FEA) simulations in ANSYS, replicating real-world wheel loading conditions.

In the simulations, nonlinear vibrations emerged in the spokes under high-frequency loading, particularly as frequency increased. To refine the model, boundary conditions were carefully set: the top of the spoke was fixed, while the bottom allowed displacement in two directions to reflect realistic deformations. A 1N, 10 Hz sine wave was applied vertically at the bottom, paired with a horizontal cosine wave of equal magnitude and frequency to simulate ground contact forces. Additionally, to simulate terrain-induced vibrations, a 0.2 N high-frequency perturbation force was superimposed, increasing linearly from 100 Hz to 500 Hz to reflect small-scale variations from rough terrain. These vibrations were crucial in capturing terrain-induced dynamic spoke behavior.

\begin{figure}[ht!] 
\centering
\includegraphics[width=2.8in]{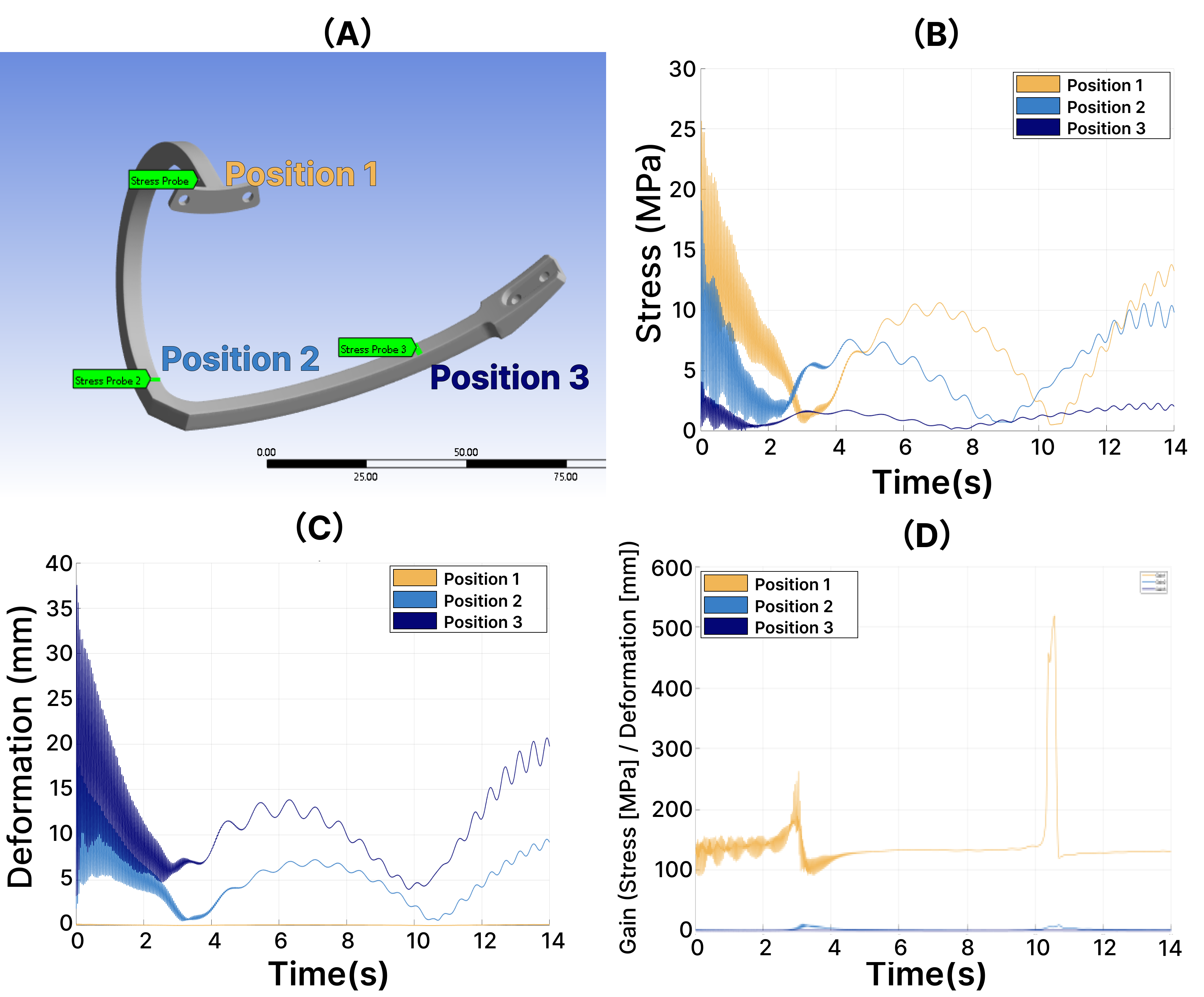}
\caption{
(A) Three measurement positions on the spoke, where stress probes are placed to record stress and deformation responses.  (B) Time series plot of stress at the three measurement positions, showing variations over time under dynamic loading.  (C) Time series plot of deformation at the three measurement positions, illustrating displacement changes due to applied forces.  (D) Time series plot of gain, defined as the ratio of stress to deformation at each position, highlighting differences in structural stiffness and response.
}
\label{Force}
\end{figure}

From the FEA results, stress distribution varied significantly depending on force direction. Horizontal forces concentrated stress at the spoke’s connection point, while vertical forces caused maximum stress at the 90-degree bend (Figure~\ref{Entropy}, B and C). These findings align with frequency-based tests, reinforcing the importance of strategic sensor placement. Three critical positions were identified on the spoke (Figure~\ref{Force}, A):

Position 1: Maximum stress under horizontal force
Position 2: Maximum stress under vertical force
Position 3: Maximum deformation under impact force

These results suggest that combining stress and vibration analysis provides a more robust terrain classification framework, leveraging the wheel’s natural mechanical response to amplify terrain-specific signals. By integrating this approach, classification accuracy can be improved in various environments, ensuring adaptability to real-world applications.

Stress and deformation data over time (Figure~\ref{Force}, B and C) reveal initial high-frequency oscillations due to large-amplitude, low-frequency forces. As the loading amplitude decreases, vibrations stabilize. However, an increasing high-frequency, low-amplitude force introduces complex dynamic behaviors.

The analysis of different positions on the spoke highlights distinct dynamic responses. Position 1 exhibits minimal vibration response and is most sensitive to low-frequency forces, making it ideal for detecting large-scale terrain deformations. Position 2 shows moderate stress variations with smooth decay, offering balanced responsiveness to both low and high frequencies. Position 3 experiences pronounced deformation vibrations and is highly responsive to high-frequency forces, making it suitable for detecting fine terrain details.

The Gain curve further highlights these dynamic differences. Gain1 peaks at 600 MPa/mm at Position 1, indicating strong stress concentration. Gain2 shows moderate fluctuations, reflecting Position 2’s role in transitioning between stability and complexity. Gain3 remains consistently low, highlighting Position 3’s stabilizing behavior under dynamic loads.

Understanding terrain-induced vibrations requires analyzing how different regions of the spoke respond dynamically to varying force frequencies. Initial stress and deformation data (Figure~\ref{Force}, B and C) reveal that at the onset, high-frequency oscillations arise due to large-amplitude, low-frequency forces. As the loading amplitude decreases, vibrations stabilize. However, the introduction of a gradually increasing high-frequency, low-amplitude force results in complex dynamic behaviors, highlighting the need for a more refined classification approach.  

To explore this further, we examined three distinct positions on the spoke, each exhibiting unique vibrational characteristics. Position 1 exhibits minimal vibration response but is highly sensitive to low-frequency forces, making it optimal for detecting large-scale terrain deformations. Position 2 shows moderate stress variations with smooth decay, providing a balanced sensitivity to both low and high frequencies. Position 3 experiences pronounced deformation vibrations and responds strongly to high-frequency forces, making it ideal for detecting fine terrain details. These differences become more evident when analyzing the Gain curve. At Position 1, Gain1 peaks at 600 MPa/mm, indicating strong stress concentration and sensitivity to structural loads. In contrast, Gain2 at Position 2 shows moderate fluctuations, reinforcing its role as a transition zone between stability and complexity. Meanwhile, Gain3 at Position 3 remains consistently low, reflecting its role in stabilizing dynamic loads.  

While Gain provides a direct measure of stress-to-deformation amplification, it does not fully capture the complexity of vibrational interactions. To quantify this, we introduce Shannon entropy, which measures the variability in the vibrational response. Higher entropy values indicate more complex, irregular force interactions, while lower entropy values suggest greater stability. To systematically evaluate these responses, we segmented stress, deformation, and gain data into probability distributions, allowing entropy to capture the variability of the system’s vibrational response. The results highlight a significant variation in Gain Shannon Entropy, ranging from 0.5797 at Position 1 to 3.1251 at Position 3, confirming that terrain-induced complexity varies significantly across different spoke locations.

\begin{figure}[ht!] 
\centering
\includegraphics[width=2.8in]{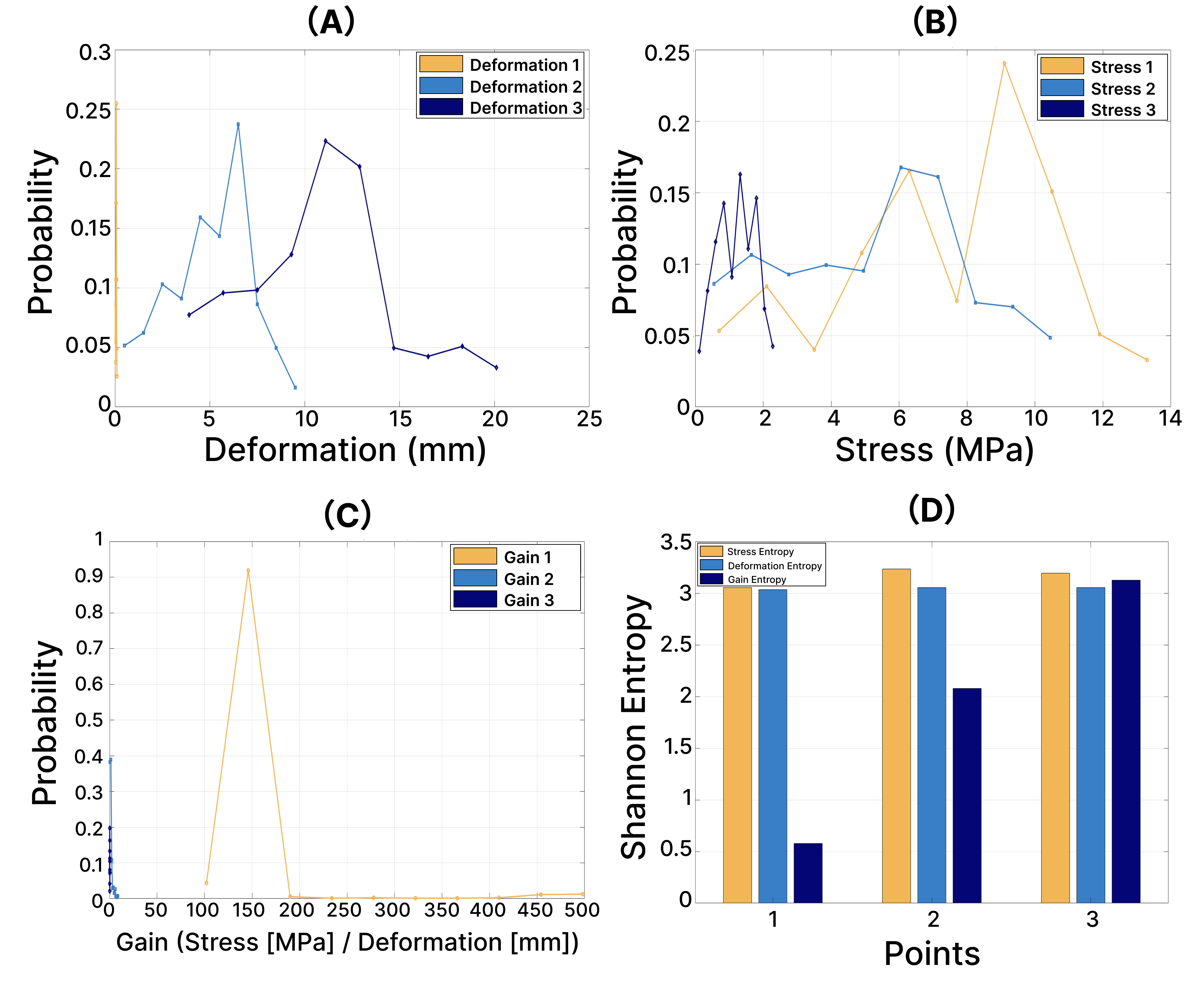}
\caption{
(A) Probability distribution of deformation at three measurement positions, showing variations in deformation magnitude. (B) Probability distribution of stress at three measurement positions, illustrating stress variability under dynamic loading. (C) Probability distribution of gain (stress-to-deformation ratio) at three positions, highlighting stiffness variations. (D) Comparison of Shannon entropy for stress, deformation, and gain across three positions, quantifying the complexity of each parameter.
}
\label{EEE}
\end{figure}

Each position exhibits distinct vibrational behaviors, reflected in their Shannon entropy values. Position 1, with an entropy of 0.5797, displays stable, predictable behavior dominated by low-frequency vibrations. The system remains largely linear and minimally influenced by high-frequency excitations or nonlinear effects. Position 2, with an entropy of 2.0769, functions as a transitional region, balancing both low- and high-frequency components and capturing a mix of stable and dynamic vibrational behaviors. Position 3, with the highest entropy of 3.1251, exhibits a highly nonlinear response, with significant high-frequency vibrational interactions, potential resonance effects, and complex system responses.  

These findings indicate a direct relationship between frequency range and vibrational behavior. At low frequencies, as seen in Position 1, the system follows predictable, linear motion, resulting in lower gain and reduced Shannon entropy. At moderate frequencies, such as in Position 2, the response balances linear and nonlinear components, yielding moderate entropy values and an intermediate vibrational profile. At high frequencies, as in Position 3, the system enters a nonlinear regime, exhibiting resonance effects, broader vibrational energy distribution, and greater sensitivity to terrain-induced variations.  

By combining FEA-based stress analysis, Gain Shannon Entropy, and vibration frequency mapping, we confirm that specific regions of the spoke are more sensitive to distinct frequency ranges. This insight is critical for optimizing sensor placement, ensuring that sensors are positioned in locations that maximize sensitivity to terrain-induced variations. Ultimately, this integration of stress and vibration analysis provides a robust framework for terrain classification, enhancing adaptability across diverse environments.

\subsection{Piezoelectric Sensor System Design and Installation}

\begin{figure}[ht!] 
\centering
\includegraphics[width=2.8in]{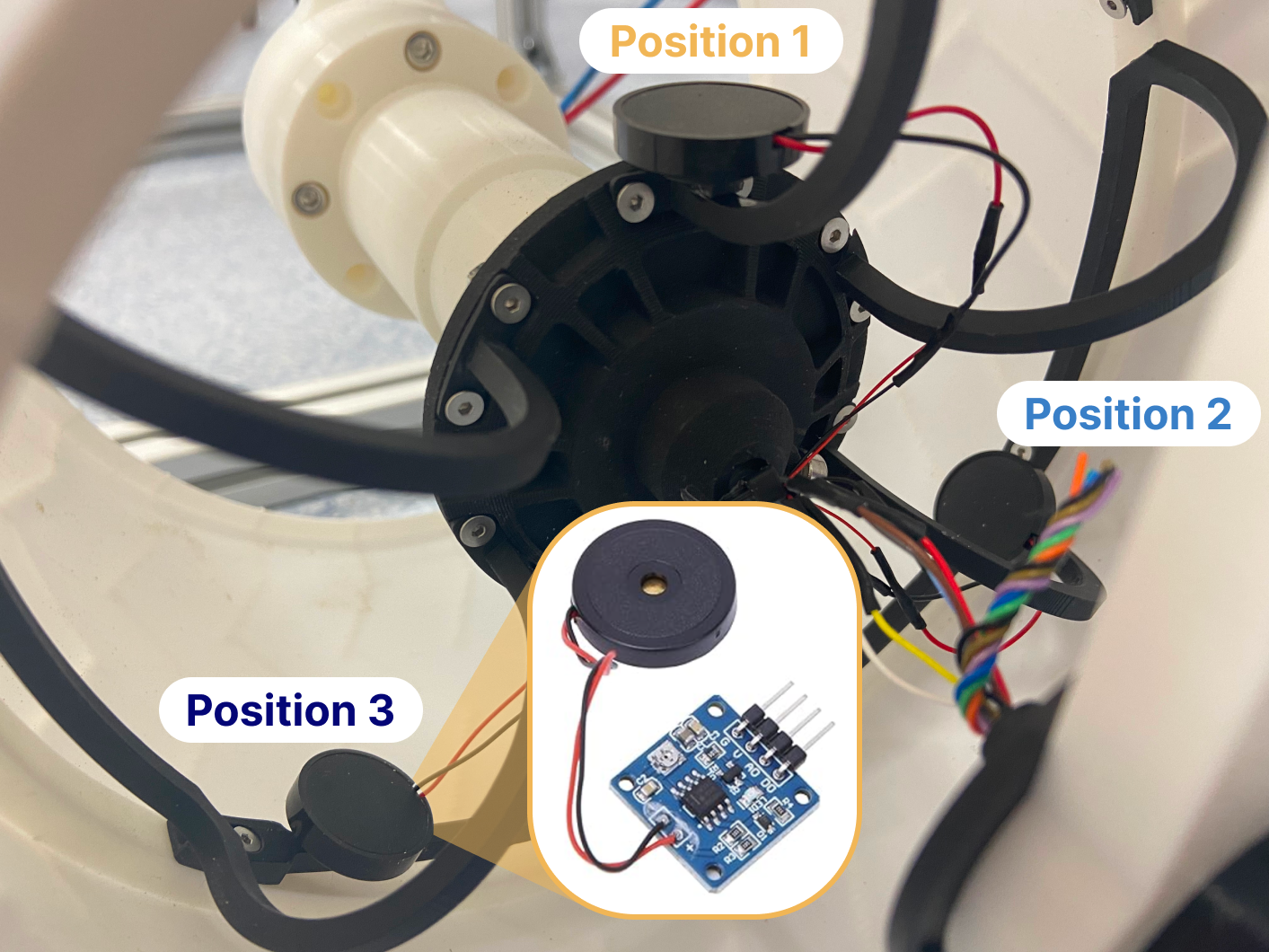}
\caption{
The placement of piezoelectric sensors within the wheel module to capture terrain-induced vibrations, leveraging the wheel’s structure to enhance sensitivity and signal differentiation.
}
\label{piezo}
\end{figure}

In this study, we used a high-sensitivity 20 mm piezoelectric vibration sensor (Figure \ref{piezo}) to capture vibrations at critical points. The sensor offers high precision, frequency adaptability, and durability, ensuring stable signal output in complex environments while maintaining ultra-low power consumption. Unlike conventional LiDAR and accelerometers, which require continuous power, the piezoelectric sensor operates passively, converting mechanical strain into electrical signals without external power. This self-powered nature significantly reduces energy consumption, making it ideal for low-power robotic applications. A low-power charge amplifier (<5 mW) enhances signal clarity while keeping overall power draw minimal.

To ensure optimal performance, a custom protective cover shields the sensor from dust and humidity, while the charge amplifier boosts weak signals for accurate low-frequency detection. The sensor was firmly installed at key vibration points, maximizing signal transfer efficiency while keeping total system power consumption below 10 mW. This energy-efficient setup enables reliable data collection and precise vibration analysis, forming a solid foundation for future terrain classification in low-power robotic systems.




\section{Experimental Method}
\label{Experimental Method}

\subsection{Rover wheel Structure Design}
To simulate the Mars rover wheel while maintaining structural consistency, we designed a 1:4 scaled model based on the "Curiosity" rover wheel. This scale preserves the structural behavior and dynamic response, making it suitable for terrain analysis. Key features, including screws and tread patterns, were replicated to ensure the model closely reflects the original wheel's characteristics.

\begin{figure}[ht!] 
\centering
\includegraphics[width=2.8in]{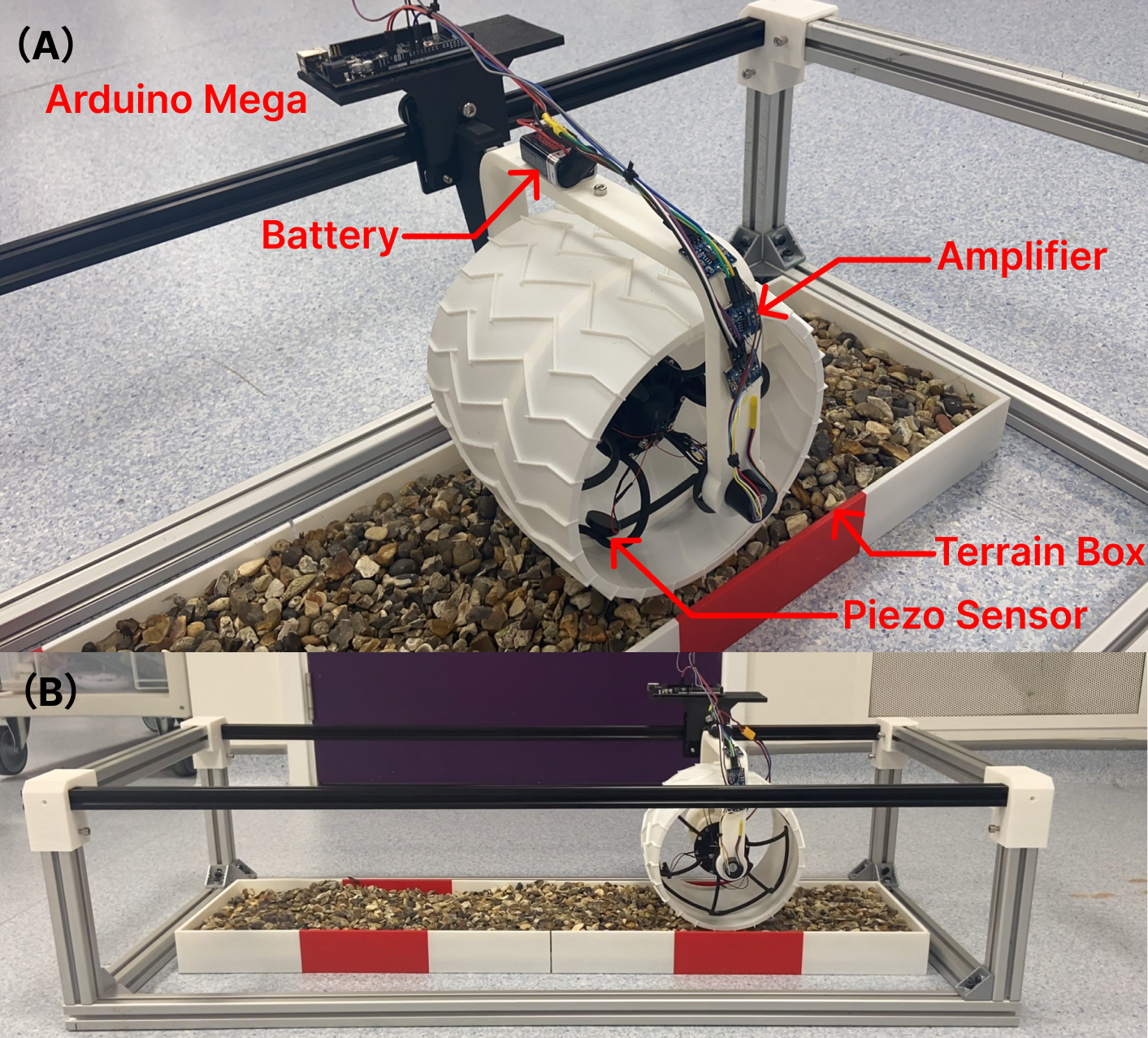}
\caption{
(A) Close-up view of the system, highlighting key components: Arduino Mega, battery, amplifier, piezoelectric sensor, and terrain box.  
(B) Full experimental setup, where the wheel moves along a controlled track over different terrain types.
}
\label{test}
\end{figure}

The wheel’s outer shell was 3D printed using Bambu PLA Basic material with a 0.2 mm layer height, balancing strength, stability, and cost. Resin 3D printing was used for the spokes to ensure a dense, solid structure, enhancing strength and stiffness for accurate Finite Element Analysis (FEA). The design, shown in Figure \ref{test}, represents a scaled-down Mars rover wheel for terrain-induced vibration analysis. A Brushed Geared DC Motor (7 W, 12 V, 1.2 Nm torque, 8 rpm) with a 721:1 gear reduction was selected to minimize vibration and provide steady operation. Its low speed aligns with NASA’s Mars rover, ensuring realistic dynamic behavior. This design effectively simulates the Mars rover wheel, supporting detailed dynamic testing and analysis.

\subsection{Data Collection} 


The test platform features an aluminum frame with a 30 cm wide and 90 cm long terrain platform. The wheel is mounted on horizontal rails for movement and a vertical rail for terrain adaptation. The motorized wheel, weighing 300g, achieves a terminal velocity of 67 mm/s, slightly exceeding Mars rover speeds. Sensors on the wheel spokes connect via slip rings to maintain wire stability during motion. The suspension system includes a piezo sensor amplifier and an Arduino for signal acquisition.

\subsection{Initial Experiment Analysis}

The initial experiments using 3 kinds of terrains: Small Stone, Sand and Flat terrain. The viberation singal is recorded, the Sampling Frequency is 720hz with total 9500 datas on each terrain using the Arduino Mega.
\begin{figure}[ht!] 
\centering
\includegraphics[width=2.8in]{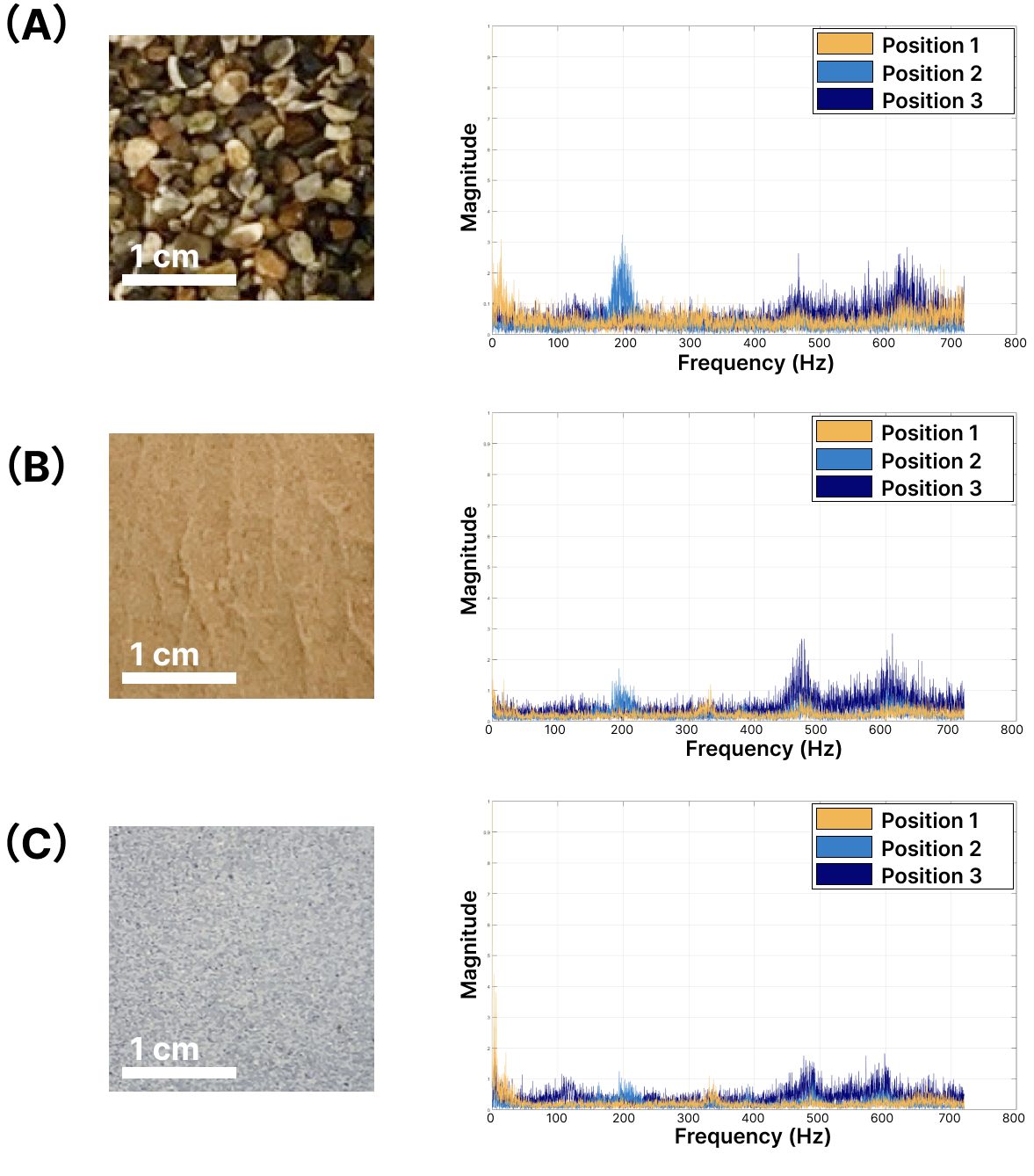}
\caption{
Three terrain surfaces and their FFT spectra, used to evaluate piezoelectric sensor performance. Clear differences appear in the low-frequency range, around 200 Hz, and higher frequencies, reflecting terrain-dependent signal variations.(A) Small rocks  (B) Sand  (C) Flat surface
}
\label{FFT}
\end{figure}

The FFT spectrum analysis of rolling vibration signals across different terrains (Figure \ref{FFT}) aligns closely with Finite Element Analysis (FEA) predictions, confirming the model's accuracy. On rocky terrain (A), irregular stone sizes produce prominent low- and mid-frequency components, with Position 2 showing a significant peak at 200 Hz. On sandy terrain (B), the dense, small sand grains generate pronounced high-frequency vibrations, especially in the 450-700 Hz range at Position 3. On flat terrain (C), the uniform surface results in minimal vibration across all frequencies. The analysis shows Position 1 is sensitive to low frequencies (0-50 Hz), Position 2 to mid-frequencies (~200 Hz), and Position 3 to high frequencies (450-700 Hz). These findings validate the FEA model and support the Shannon Entropy analysis, demonstrating that sensors at specific positions effectively capture terrain features within designated frequency bands. The distinct frequency characteristics across terrains underscore the potential for accurate classification based on vibration signals.

\begin{figure}[ht!] 
\centering
\includegraphics[width=2.8in]{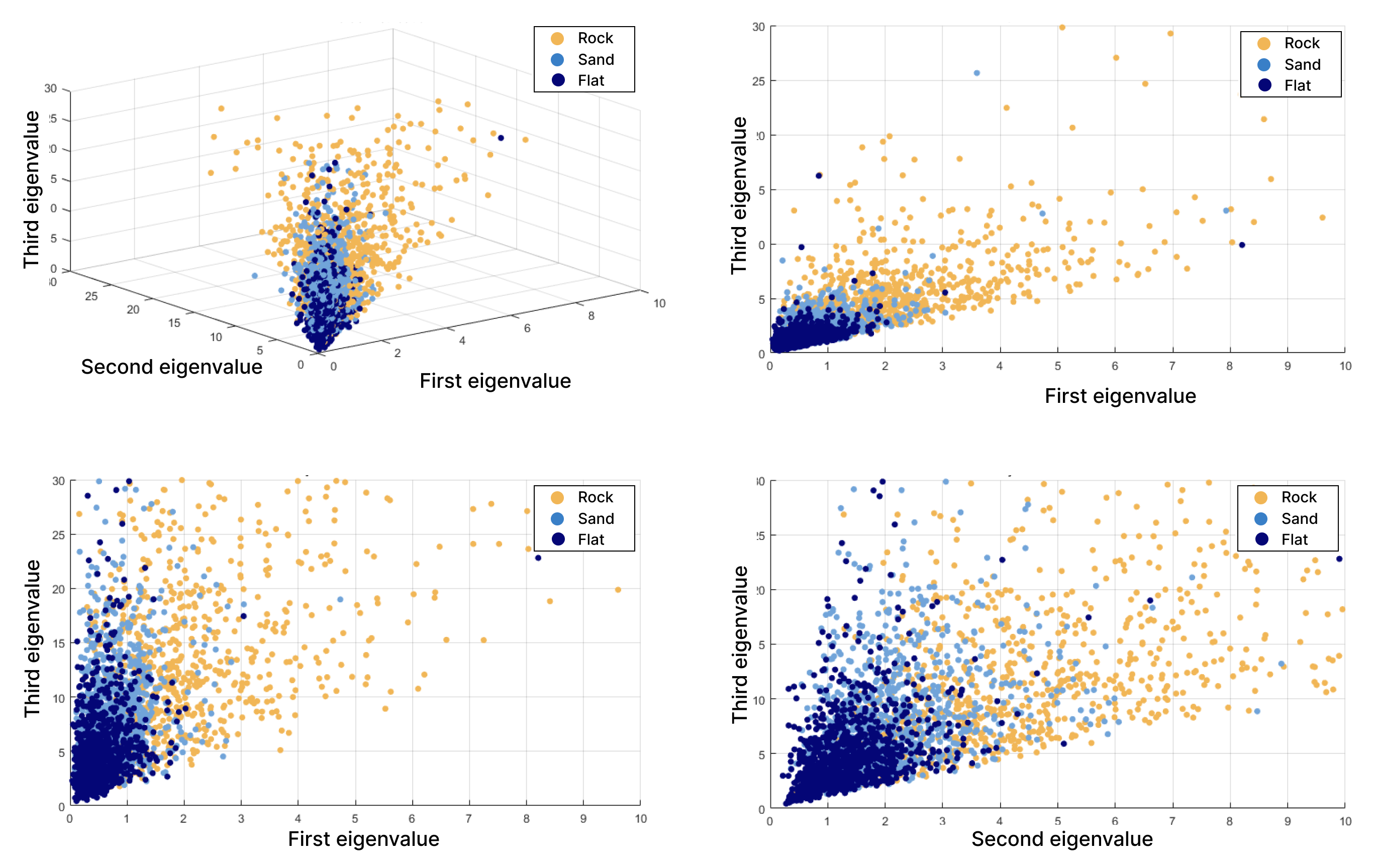}
\caption{
Eigenvalue distributions of piezoelectric sensor outputs for three terrain surfaces, mapped in eigen-space for terrain classification. The plots illustrate the separation of terrain types (rock, sand, and flat) based on the first, second, and third eigenvalues, highlighting their distinct clustering patterns.
}
\label{eigen}
\end{figure}
The eigenvalue distributions in Figure \ref{eigen} clearly distinguish different terrains. Computed from the 3×3 covariance matrix of piezo sensor outputs at different wheel positions, these eigenvalues capture signal variance and inter dependencies, reflecting terrain-induced vibration propagation.

Rocky terrain exhibits large, dispersed eigenvalues, indicating high variance from its irregular surface. Sand terrain shows moderate eigenvalue spread, reflecting its semi-uniform nature. Flat terrain has small, tightly clustered eigenvalues, representing minimal variance and a stable surface.

The eigenvalue spread highlights terrain complexity: rocky terrain has the highest variance, flat the lowest, and sand lies in between. The distinct clusters, especially in 2D projections, support machine learning-based classification (e.g., k-NN, SVM). Additionally, eigenvalues correlate with vibration energy levels, where larger values indicate dynamic interactions (rocks), smaller values suggest stability (flat terrain), and sand exhibits intermediate characteristics.

\subsection{Feature Extraction}
Building on the initial analysis highlighting distinct frequency characteristics across terrains, the next step involves feature extraction and machine learning for terrain classification. Features capturing vibrational responses across frequency bands were extracted from sensors at three wheel-spoke positions, each sensitive to specific frequencies: Position 1 to low, Position 2 to mid, and Position 3 to high frequencies. These features, optimized for terrain classification, were used as inputs for the machine learning model, detailed in the following sections.

\subsubsection{\textbf{Position 1 Sensor Features}}
Position 1, sensitive to low frequencies, focused on features in the 1–50 Hz range:

\textbf{Low-Frequency RMS}: Calculated to measure the energy of the low-frequency signal, representing terrain compliance. Lower RMS values indicate softer terrains like sand, while higher values suggest harder surfaces like rocks.

\textbf{Low-Frequency Standard Deviation}: Quantifies variability in low-frequency signals to assess terrain uniformity where higher values suggest significant variability, often linked to rocky terrains.

\subsubsection{\textbf{Position 2 Sensor Features}}
Position 2 primarily captured mid-frequency vibrations, focusing on periodicity and vibrational consistency:

\textbf{Autocorrelation Peak}: Measures signal periodicity, helping identify regular terrain patterns such as evenly spaced stones which prominent peaks indicate repeating terrain features.

\textbf{Smoothness of Amplitude Variation}: Assesses how smoothly signal amplitudes change, indicating terrain transitions where higher values correspond to smoother terrains like sand.

\subsubsection{\textbf{Position 3 Sensor Features}}
Position 3, sensitive to high frequencies (400–800 Hz), targeted fine surface irregularities:

\textbf{High-Frequency Filtered Standard Deviation}: Measures variability in high-frequency signals to identify surface roughness where higher values indicate rough, uneven terrains like cobblestones.

\textbf{Kurtosis (Spike Rate)}: Identifies sharp peaks in the signal, representing terrains with significant irregularities where higher kurtosis values highlight highly uneven terrains with frequent obstacles.

By tailoring feature extraction to each sensor's specific frequency sensitivity, the data captures a diverse range of terrain characteristics, providing a robust basis for machine learning-based terrain classification

\section{RESULTS}
\label{RESULTS}
\begin{figure}[ht!] 
\centering
\includegraphics[width=2.8in]{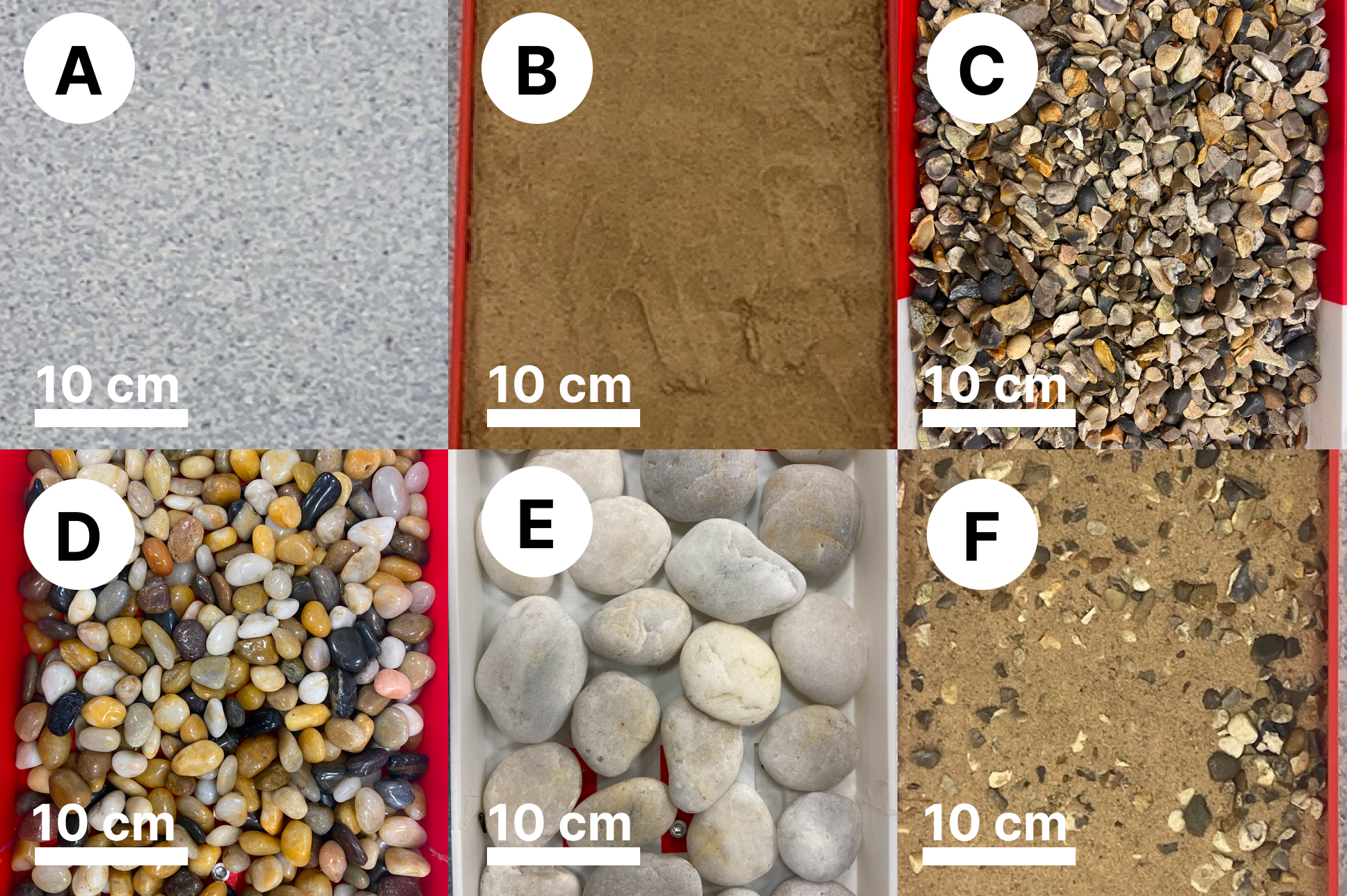}
\caption{
Six terrains used in the machine learning and the unknown terrain classification, (A)Flat terrain, (B)Fine sandy terrain (<1mm), (C)Small Stone terrain (5mm - 10mm), (D)Small Pebble terrain (20mm - 30mm), (E)Large Stone terrain (50mm - 70mm), (F)Unknown terrain (mixture of the sand and small stone).
}
\label{six}
\end{figure}

To validate the proposed methods, terrain identification experiments were conducted using piezoelectric sensors mounted on the rover wheel spokes across multiple terrain types as shown in Figure\ref{six}. Vibrational data were analyzed based on features extracted from low, mid, and high-frequency ranges specific to each sensor position.

\subsection{SVM Classification Results}

The SVM classifier was used to identify five distinct terrain types, as shown in Figure \ref{six} from A to E, based on features extracted from vibrational responses recorded by piezoelectric sensors. The extracted features included root mean square (RMS), standard deviation, kurtosis, skewness, signal energy, and entropy, corresponding to equations (1-6). These features were computed separately for low-, mid-, and high-frequency bands, aligning with sensor placements on the rover wheel spokes to capture unique vibrational characteristics for each terrain.

\begin{figure}[ht!] 
\centering
\includegraphics[width=2.8in]{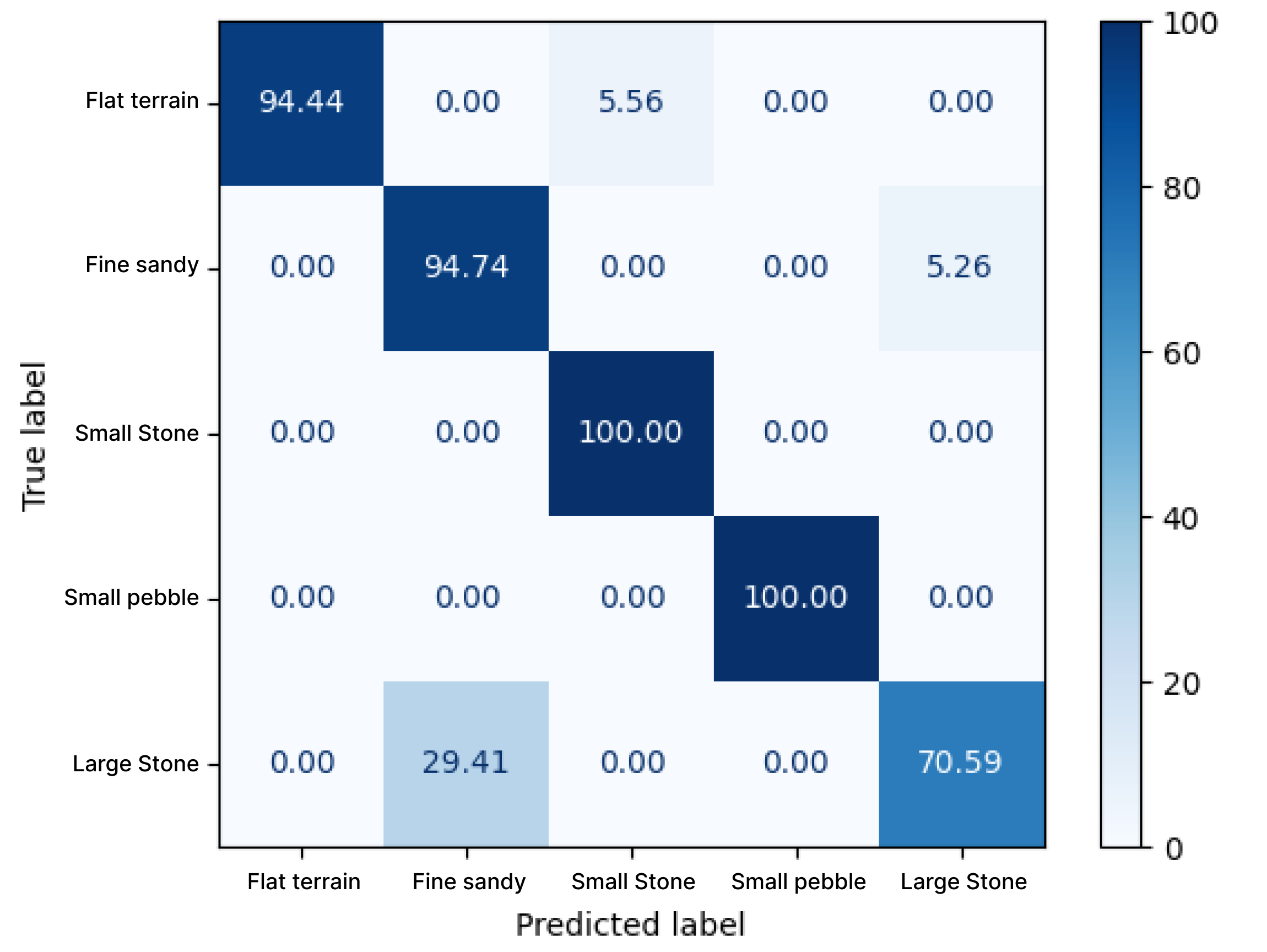}
\caption{
Confusion matrix of reservoir computing prediction success rate averaged over 120 randomly trials (2000 data sets), with robot speed of 67 mm/s and the sampling frequency of the whisker at 1440 Hz with the final accuracy is 90\%.
}
\label{ConfusionMatrix}
\end{figure}

When using a 1.5-second time window with 2000 rows of data, the classifier achieved 90\% accuracy, as shown in Figure~\ref{ConfusionMatrix}. The high sampling frequency provided sufficient data despite the short time frame, enabling the wheel vibrations to identify terrain and stone types quickly. The confusion matrix reveals clear distinctions between materials, such as small stones, small pebbles, and flat ground, with some classifications reaching 100\% accuracy. However, lower accuracy was observed between large stones and sand, likely due to slipping on large stones, which increased vibration frequencies and caused feature overlap.

Overall, the high accuracy, despite using only six statistical features, highlights the efficiency and reliability of the SVM model for terrain classification. The combination of frequency-domain feature extraction and machine learning classification demonstrates the feasibility of using minimal computational resources for rapid and effective terrain identification.

\subsection{Surface Roughness Estimation of Unknown Terrains}
Understanding surface roughness is essential for predicting unknown terrains, as different terrains induce distinct vibration characteristics. This study estimates terrain roughness by extracting features from high-, mid-, and low-frequency signal bands, including RMS, standard deviation, and kurtosis, to characterize terrain-induced vibrations. Known terrains are averaged into a representative feature matrix, while unknown terrains are processed to generate feature vectors for comparison.

To quantify terrain similarity, we employ two complementary distance metrics: Euclidean distance, which measures geometric proximity, and Mahalanobis distance, which accounts for statistical dependencies between features. By combining both, we achieve a more comprehensive terrain classification approach.

While Euclidean distance provides a straightforward similarity measure by computing the direct distance between feature vectors, it assumes independent and equally weighted features. However, real-world terrain vibration data often exhibits correlations, meaning that variations in one feature can influence others. Ignoring these dependencies may result in misleading classifications, particularly for complex terrains.

To address this, Mahalanobis distance is introduced, leveraging the covariance structure of the known terrain feature space to normalize feature contributions based on their variability. This prevents highly correlated features from dominating similarity calculations, resulting in a more robust and reliable classification system.

Euclidean distance directly reflects the overall magnitude difference between feature vectors, making it computationally efficient and effective when feature scales are uniform. In contrast, Mahalanobis distance is formulated as:
\begin{equation}
    d_M(x, y) = \sqrt{(x - y)^T S^{-1} (x - y)},
\end{equation}
where $x$ and $y$ are the feature vectors of two terrains, $S$ is the covariance matrix of the known terrain features, and $S^{-1}$ accounts for feature correlations. By adjusting distances based on covariance relationships, Mahalanobis distance effectively weights feature contributions, preventing dominant features from overshadowing others. Regularization is applied to the covariance matrix to maintain numerical stability.

\begin{figure}[ht!] 
\centering
\includegraphics[width=2.8in]{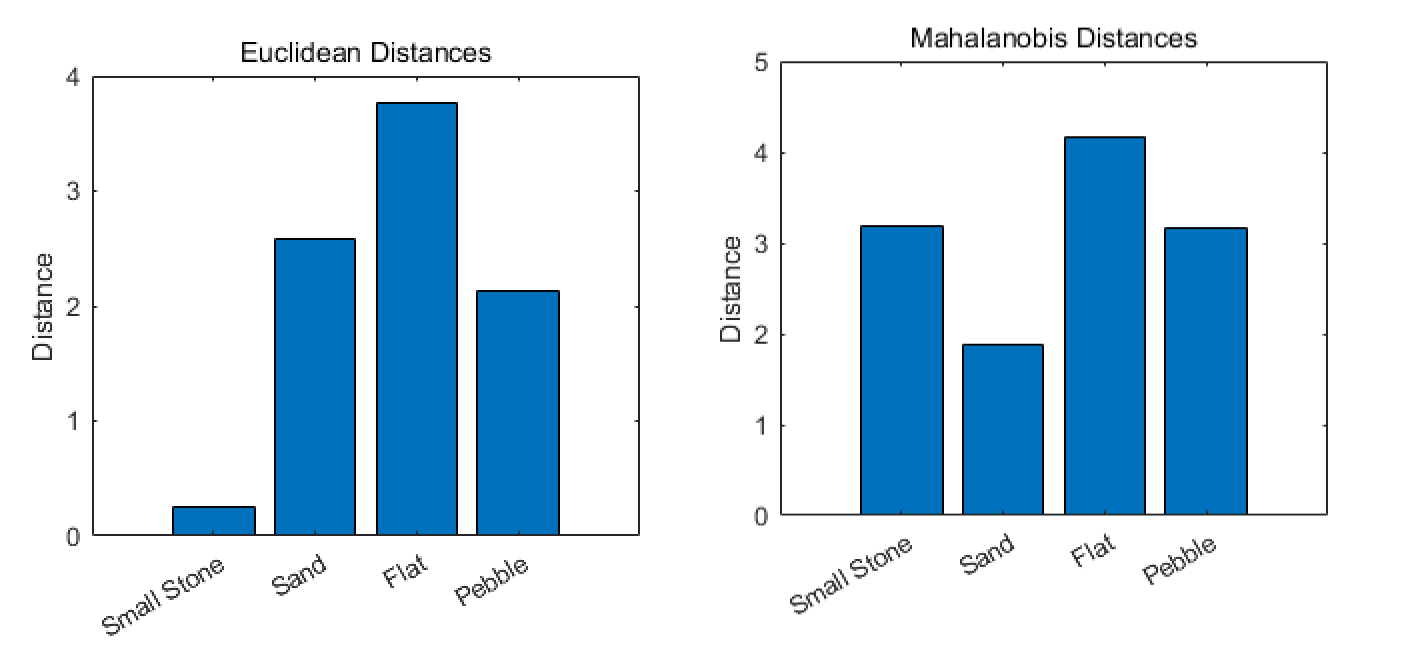}
\caption{
(A) The Euclidean distances between the unknown terrain and each known terrain, where a smaller value indicates closer geometric proximity in feature space.
(B) The Mahalanobis distances, where smaller values indicate closer alignment in the statistical feature distribution.
}
\label{predict}
\end{figure}

Figure \ref{predict} illustrates the computed distances for unknown terrain classification. Based on Euclidean distance, the unknown terrain is closest to Small Stone (value: $0.2535$), indicating geometric similarity. However, Mahalanobis distance suggests greater alignment with Sand (value: $1.8759$), reflecting statistical similarity.

This divergence suggests that the unknown terrain shares both geometric and statistical characteristics with Small Stone and Sand, highlighting the importance of using both metrics together. The Euclidean distance captures shape-based similarities, whereas Mahalanobis distance accounts for variability and feature interdependencies. 

By combining Euclidean and Mahalanobis distances, this approach enhances the robustness of real-time unknown terrain estimation, ensuring that both direct feature magnitudes and statistical dependencies are considered for a more accurate classification.

\section{CONCLUSIONS}
This study demonstrates that the hook-shaped spoke of a rover wheel can function as a mechanical reservoir, effectively separating terrain-related frequency components in time domain directly.  By leveraging this property, terrain identification is simplified using just three piezoelectric sensors strategically placed along the spoke.  The optimal sensor placement is determined through Finite Element Analysis (FEA) and Shannon entropy-based frequency localization, ensuring maximum sensitivity to terrain-induced vibrations. Moreover, combining vibration and stress analysis improves classification accuracy compared to vibration-based methods alone, confirming the advantages of a multimodal sensing approach. Experimental results show that with three sensors, the system achieves a 90\% classification accuracy. Future research will focus on improving system robustness in dynamic outdoor conditions, integrating machine learning for adaptive feature selection, and exploring variable-stiffness piezoelectric sensors to enhance sensitivity and adaptability.  These advancements will further support energy-efficient, vibration-driven terrain classification, contributing to the development of terrain-aware robotics for planetary rovers and autonomous ground vehicles.

\addtolength{\textheight}{-12cm}   




\bibliography{main} 

\end{document}